# WAVELET COHERENCE OF TOTAL SOLAR IRRADIANCE AND ATLANTIC CLIMATE

VASIL KOLEV[1], YAVOR CHAPANOV[2],

[1]*Institute of Information and Communication Technologies, Bulgarian Academy of Sciences*
*Acad. Georgi Bonchev Str., Block 2, Sofia 1113, Bulgaria*
[2]*Climate, Atmosphere and Water Research Institute, Bulgarian Academy of Sciences*
*Acad. Georgi Bonchev Str., Block 6, Sofia 1113, Bulgaria*
E-mail: kolev_acad@abv.bg, vasil.kolev@iict.bas.bg
E-mail: yavor.chapanov@gmail.com

**DEDICATED TO THE MEMORY OF PROF. MILCHO TSVETKOV**

**Abstract.**  The oscillations of climatic parameters of North Atlantic Ocean play important role in various events in North America and Europe. Several climatic indices are associated with these oscillations. The long term Atlantic temperature anomalies are described by the Atlantic Multidecadal Oscillation (AMO). The Atlantic Multidecadal Oscillation also known as Atlantic Multidecadal Variability (AMV), is the variability of the sea surface temperature (SST) of the North Atlantic Ocean at the timescale of several decades. The AMO is correlated to air temperatures and rainfall over much of the Northern Hemisphere, in particular in the summer climate in North America and Europe. The long-term variations of surface temperature are driven mainly by the cycles of solar activity, represented by the variations of the Total Solar Irradiance (TSI). The frequency and amplitude dependences between the TSI and AMO are analyzed by wavelet coherence of millennial time series since 800 AD till now. The results of wavelet coherence are compared with the detected common solar and climate cycles in narrow frequency bands by the method of Partial Fourier Approximation. The long-term coherence between TSI and AMO can help to understand better the recent climate change and can improve the long term forecast.

# 1. INTRODUCTION

Many studies have shown coherence between TSI and other important time series parameters (Benevolenskaya et al., 2013). In (Chapanov, 2021) the narrow frequency bands of the cycles of Atlantic multidecadal oscillation (AMO) are compared with corresponding oscillations of total Solar Irradiance (TSI). By using the method of Partial Fourier Approximations (PFA) between interannual and decadal cycles, the correlation between the AMO and the TSI are shown for periodicities between 48 and 233 years. The correlation between TSI and AMO is also considered in (Velasco et al., 2008; Swingedouw et al., 2010; Shindell et al., 2001).

Wavelet transforms are one well-known approach for time series processing. Wavelets are basis functions possess the important properties – orthogonality, compact support, optimal time and frequency resolution, and so on. When used in time series analysis wavelet transforms capture the non-stationarity in the spectrum very well. The analysis of different type signals has the ability simultaneously to extract local spectral and temporal information by decomposition of different resolutions. Therefore, they provide a space-scale or time-frequency analysis, they are useful a tool for the investigation of time series.

The continuous wavelet transform (CWT) allows for the analysis of non-stationary signals at multiple scales and can analyze a signal with good resolution in both time and frequency by using continuous wavelets (for example - Morlet, Paul, DoG, Mexican Hat). The obtained wavelet coefficients are functions of the scale and positions as they are computed as sums over all time of the signal multiplied by scaled and shifted versions of the wavelet. In time series analysis using the CWT the continuous Morlet wavelet is preferred because it has optimal joint time-frequency concentration. The mathematics of the CWT was first introduced in (Goupillaud et al., 1984). In the time-frequency analysis it has advantages because it is Gaussian-shaped in the frequency domain and can retain the temporal resolution of the signal. Its convolution is implemented by using the FFT which is computationally efficient.  By using continuous wavelet (Xu et al., 2021) is construct the TSI reconstruction model and the monthly total solar irradiance (TSI) time series for 1907 to 1978 years. The results of CWT for the period from 1979 to 2015 years show that the TSI and sunspot number all have significant and stable oscillation periods of 9~13 years and intermittent oscillation periods of 2~6 months only during the times of intense solar activity.

In this paper, we focus on the TSI reconstruction based on sunspot number and AMO index. First, we consider in sections 1.1, 1.2, and 2 the two methods for processing time series – the partial Fourier approximation and the estimation of the wavelet spectrum, and their coherence. Second, in sections 3 and 4 the TSI and AMO are introduced in brief. Finally, in section 5 we study how to examine the solar periodicities that could possibly be related with the AMO for the time period 800 – 2010 (Wang et al., 2017) and the coherence with the TSI for the time period 850 – 2010 (Fröhlich et al, 2018).

## 1.1 PARTIAL FOURIER APPROXIMATION

The periodic variations of time series are determined by the Method of Partial Fourier Approximation (PFA). The time series of oscillations from a given frequency band are calculated as a superposition of two neighbor Fourier harmonics, whose coefficients are estimated by the Least Squares (LS) Method. The details of this method are described in (Chapanov et al., 2015). Shortly, the partial Fourier approximation $F(t)$ of discrete data is given by:

$$F(t) = f_0 + f_1(t - t_0) + \sum_{k=1}^{n} a_k \sin \frac{2\pi k}{P_0}(t - t_0) + \sum_{k=1}^{n} b_k \cos \frac{2\pi k}{P_0}(t - t_0) \quad (1)$$

where $P_0$ is the period of the first harmonic, $t_0$ - the mean epoch of observations, $f_0$, $f_1$, $a_k$ and $b_k$ are unknown coefficients and $n$ is the number of harmonics of the partial sum, which covers all oscillations with periods between $P_0/n$ and $P_0$. The application of the LS estimation of Fourier coefficients needs at least $2n + 2$ observations, so the number of harmonics $n$ is chosen significantly smaller than the number $N$ of sampled data $f_i$. The small number of harmonics $n$ yields to LS estimation of the coefficient errors. This estimation is important difference with the classical Fourier approximation. The other difference is the possibility of arbitrary choice of the period of first harmonic $P_0$, instead of the observational time span, so the estimated frequencies may cover the desired set of real oscillations. This method allows a flexible and easy separation of harmonic oscillations into different frequency bands by the formula:

$$B(t) = \sum_{k=m_1}^{m_2} a_k \sin k \frac{2\pi}{P_0}(t - t_0) + \sum_{k=1}^{n} b_k \cos k \frac{2\pi}{P_0}(t - t_0) \quad (2)$$

where the desired frequencies $\omega_k$ are limited by the bandwidth

$$\frac{2\pi m_1}{P_0} \leq \omega_k \leq \frac{2\pi m_2}{P_0}, \quad (3)$$

After estimating the Fourier coefficients, it is possible to identify a narrow frequency band presenting significant amplitude, and defining a given cycle. Then this cycle can be reconstructed in time domain as the partial sum limited to the corresponding frequency bandwidth.

## 1.2 CONTINUOS WAVELET TRANSFORM

The Morlet wavelets are a one – parameter family of functions and consist of a complex exponential modulated by a Gaussian (Cohen, 2019),

$$\psi(t) = e^{i2\pi f_0 t} e^{-\frac{4\ln(2)t^2}{h^2}} \quad (4)$$

where $h$ - is full-width at half-maximum (FWHM), and in seconds is the distance in time between 50% gain before the peak to 50% gain after the peak. Therefore, for a given square – integrable signal $s(t) \in L^2(R)$ in the time domain $(-\infty, \infty)$, analyzing with the continuous wavelet transform (CWT) is the inner product of the signal with a series of daughter wavelet functions and is defined with an equation, i.e.

$$W(a,b) = <s,\psi> = \frac{1}{\sqrt{a}} \int_{-\infty}^{\infty} s(t)\psi(\frac{t-b}{a})dt, \quad (5)$$

where $L^2(R)$ denotes the Hilbert space of measurable, square – integrable one-dimensional signals, $\psi(t)$ represents a mother wavelet, $a$ is a dilation/compression scale factor that determines the characteristic frequency, $b$ - the translation. Therefore, by reducing the scaling parameter $a$ we can reduce the support of the wavelet in space, which covers higher frequencies, and vice versa. This means that $1/a$ is a measure of frequency. The parameter $b$ indicates the location of the wavelet window along the space axis.

Wavelet dilation increases the CWT's sensitivity to long time-scale events, and wavelet contraction increases its sensitivity to short time-scale events. This changing of the pair ($a$, $b$) enables computation of the wavelet coefficients $W(a, b)$ on the entire space-frequency plane. Reconstruction of a signal from the wavelet coefficients is obtained by the expression:

$$s(t) = \frac{1}{C_\psi} \int_{-\infty}^{0} \int_{0}^{\infty} W(a,b)|a|^{-\frac{1}{2}} \psi\left(\frac{t-b}{a}\right) \frac{dadb}{a^2} \quad (6)$$

where the Fourier transform of the mother wavelet $\hat{\psi}(\omega)$ needs to satisfy the following admissibility condition,

$$C_\psi = \int_{-\infty}^{\infty} \frac{|\hat{\psi}(\omega)|^2}{|\omega|} d\omega < +\infty,$$

where $C_\psi$ is the *admissibility constant* depending on the chosen wavelet. It follows that $\hat{\psi}(\omega)$ is a continuous function, so that the finiteness of $C_\psi$ implies that $\hat{\psi}(0) = 0$, has no zero frequency component and the mean value of $\psi(t)$ in the time domain is zero, i.e. $\int_{-\infty}^{\infty} \psi(t)dt = 0$. The wavelet $\psi(t)$ is a window function that simultaneously enables the possibility of time–frequency localization and has finite energy

$$E = \int_{-\infty}^{\infty} |\psi(t)|dt < +\infty.$$

More details for the CWT are given in (Sadowsky, 1994, 1996; Najmi et al., 1997). The absolute values of the CWT show a function of time and frequency that is *a scalogram*. Since the "*effective support*" of the wavelet at scale $a$ is proportional to $a$, these edge – effects also increase with $a$. The region in which the transform suffers from these edge effects is called cone of influence (COI). COI is the set of all $t$ included in the effective support $supp_{eff}(b-aK, b+aK)$ of the wavelet at a given position and scale. At each scale, the COI determines the set of wavelet coefficients influenced by the value of the signal at a specified position. Usually, the COI includes the line and the shaded region from the edge of the line to the frequency (or period) and time axes as it shows areas in the scalogram potentially affected by *edge-effect artifacts* (Mallat, 1999). It is important to note that these effects arise from areas where the stretched wavelets extend beyond the edges of the observation interval. Therefore, the information provided by the scalogram is an accurate time-frequency representation of the data within the unshaded region delineated by the white line, while outside the white line in the shaded region, information in the scalogram can be treated as suspect.

## 2. WAVELET COHERENCE AND PHASE

The wavelet coherence between time series is related to the periodic phenomena in time series. It is similar to a bivariate correlation coefficient and measures the degree and extent of co-movement between two-time series variables but in the time-frequency domain. A wavelet coherency could identify both frequency bands and time intervals where the time series relate. It is defined using smoothing in both time and scale, with the amount of smoothing dependent on both the choice of wavelet and the scale. Given two time series *X* and *Y*, with wavelet transforms $W_n^X(s)$ and $W_n^Y(s)$ the cross-wavelet spectrum is defined as

$$W_n^{XY}(s) = W_n^X(s) W_n^{Y*}(s) \qquad (7)$$

where *n* is the time index, *s* is the scale, and (*) indicates complex conjugate. The *wavelet-squared coherency* is defined as the absolute value squared of the smoothed cross-wavelet spectrum, normalized by the individual wavelet power spectra (Velasco et al., 2008),

$$R_n^2(s) = \frac{S(s^{-1}|W_n^{XY}(s)|^2)}{S\langle s^{-1}|W_n^X(s)|^2\rangle S(s^{-1}|W_n^Y(s)|^2)}, \ 0 \leq R_n^2(s) \leq 1 \qquad (8)$$

where $\langle \cdot \rangle$ indicates smoothing in both time and scale, *S* is a smoothing operator. The factor $s^{-1}$ is used to convert to an energy density. Finally, it is noted that because the wavelet transform conserves variance, the wavelet coherency is an accurate representation of the (normalized) covariance between the two time series.

The wavelet-coherency phase difference given by

$$\phi_n(s) = \tan^{-1}\left(\frac{\text{Re}(s^{-1}W_n^{XY}(s))}{\text{Im}(s^{-1}W_n^{XY}(s))}\right) \qquad (9)$$

where the smoothed real and imaginary parts is already been calculated in (8). Both $R_n^2(s)$ and $\phi_n(s)$ are functions of the time index *n* and the scale *s*. The smoothing in (8) and (9) was been done using a weighted running average (or convolution) in both the time and scale directions.

It should be note that wavelet coherency depends on an arbitrary smoothing function. Nevertheless, a ''natural'' width of the smoothing function in both time and Fourier space can be provides use the Morlet wavelet function. The time smoothing uses a filter given by the absolute value of the wavelet function at each scale, normalized to have a total weight of unity. The significance level of the wavelet coherence is also determined using the Monte Carlo simulations.

The advantage of wavelet coherence over wavelet power is that it shows statistical significance only in areas where the series involved actually share significant periods. Its disadvantage may be that it needs careful fine-tuning in computation and plotting.

### 3. TOTAL SOLAR IRRADIANCE

The TSI (Fligge et al., 2000, Foukal, 2012) is the value of the integrated solar energy flux over the entire spectrum arriving at the top of the terrestrial atmosphere at the mean Sun - Earth distance (the astronomical unit, AU). Mathematical calculations, as well as satellite observations indicate an average value of 1367 ± 4 W/m$^2$. After distribution of the solar irradiance over the planetary atmosphere, the average solar radiation at the top of the atmosphere is 1/4 of this: 342 W/m$^2$. The Earth's planetary albedo of 0.3 further reduces the incoming radiation to 239 W/m$^2$. Upon entering the atmosphere solar irradiance wavelengths shorter than 300 nm are absorbed in the stratosphere and above. Satellite measurements of the TSI started with NIMBUS-7 launched in November 1978 and has been carried out by their successors.

The TSI has been widely used as a characteristic of solar variability as its variations can be separate into a cyclic and a long-term component that is calculated using sunspots and faculae as the main contributors.

### 4. ATLANTIC MULTIDECADAL OSCILATION (AMO) INDEX

The AMO (Enfield et al., 2001) is ocean oscillations system which sets the character of regional changes in oceanic environment and is the theorized variability of the sea surface temperature (SST) or a climate mode in the North Atlantic Ocean (NAO) from 0° to 70° N, occurring at the time scale of several decades. Its index is a measure of basinwide sea surface temperature variation in the North Atlantic, adjusted to remove trends in anthropogenically forced warming.

Like other modes of variability (e.g., El Niño Southern Oscillation), the AMO has impacts on a large geographic scale via atmospheric teleconnections, and has been hypothesized to have an influence on a range of North Atlantic fisheries and ecosystems. AMO has been identified as a coherent mode of natural variability occurring in the North Atlantic Ocean with an estimated period of 60-80 years. It is based upon the average anomalies of sea surface temperatures (SST) in the North Atlantic basin, typically over 0-80N. To remove the climate change signal from the AMO index, users typically detrend the SST data at each grid point or detrend the spatially averaged timeseries.

## 5. RESULTS

### 5.1 SOLAR INFLUENCE ON CENTENNIAL AND DECADAL TSI AND AMO VARIATIONS

All solar influences of TSI and AMO variations estimated by using of partial Fourier approximation are obtained in (Chapanov, 2021). The correlation periodicities between AMO and TSI for 850 – 2011 are presented in Table 1. The agreement between time series with these oscillations is excellent with a high correlation between them for the time interval 1100-2010 yrs.

**Table 1.** The correlation periodicities for 850 – 2011 years between AMO and TSI

|   | Correlations | |
|---|---|---|
|   | Decadal, years | Centennial, years |
| 1 | 72.4-77.3 | 193.2-232.8 |
| 2 | 64.4-68.2 | 144.9-165.6 |
| 3 | 58.0-61.0 | 115.9-128.8 |
| 4 | 48.3-54.4 |   |

### 5.2 CWT OF TSI AND AMO

The used time series of TSI and AMO long-term variations cover 1161years and 1211 years overlapped time interval for the period 850.0 – 2011 yrs and 800.0 – 2011 yrs, respectively shown of fig.1. The power spectra of the individual AMO and TSI time series with CWT using the Morlet wavelet is shown in fig. 2. The vertical and horizontal axes show the scale (years) bands and a period in years, respectively. The significant co-movements are marked by dark red colors on the top of the sidebar. In contrast, colder blue colors signify lower dependence between the series. Cold regions beyond the significant areas represent time and frequencies with no dependence in the series. The thin black line and the gray line cone delineate the cone of influence (COI) region where the estimates of wavelet coefficients are statistically insignificant at a 5% level of significance. This is estimate by Monte Carlo simulation with phase-randomized surrogate series. Lightly shaded zones inside the COI are ruled out from the analysis, due to the potential presence of edge artifacts.

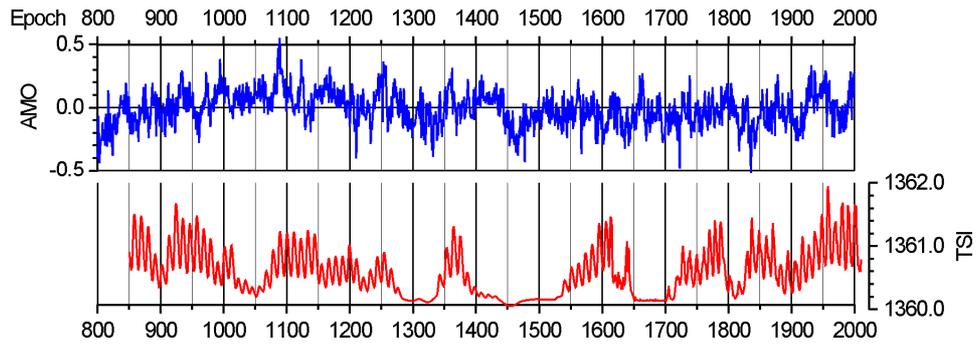

Figure 1. AMO and TSI data

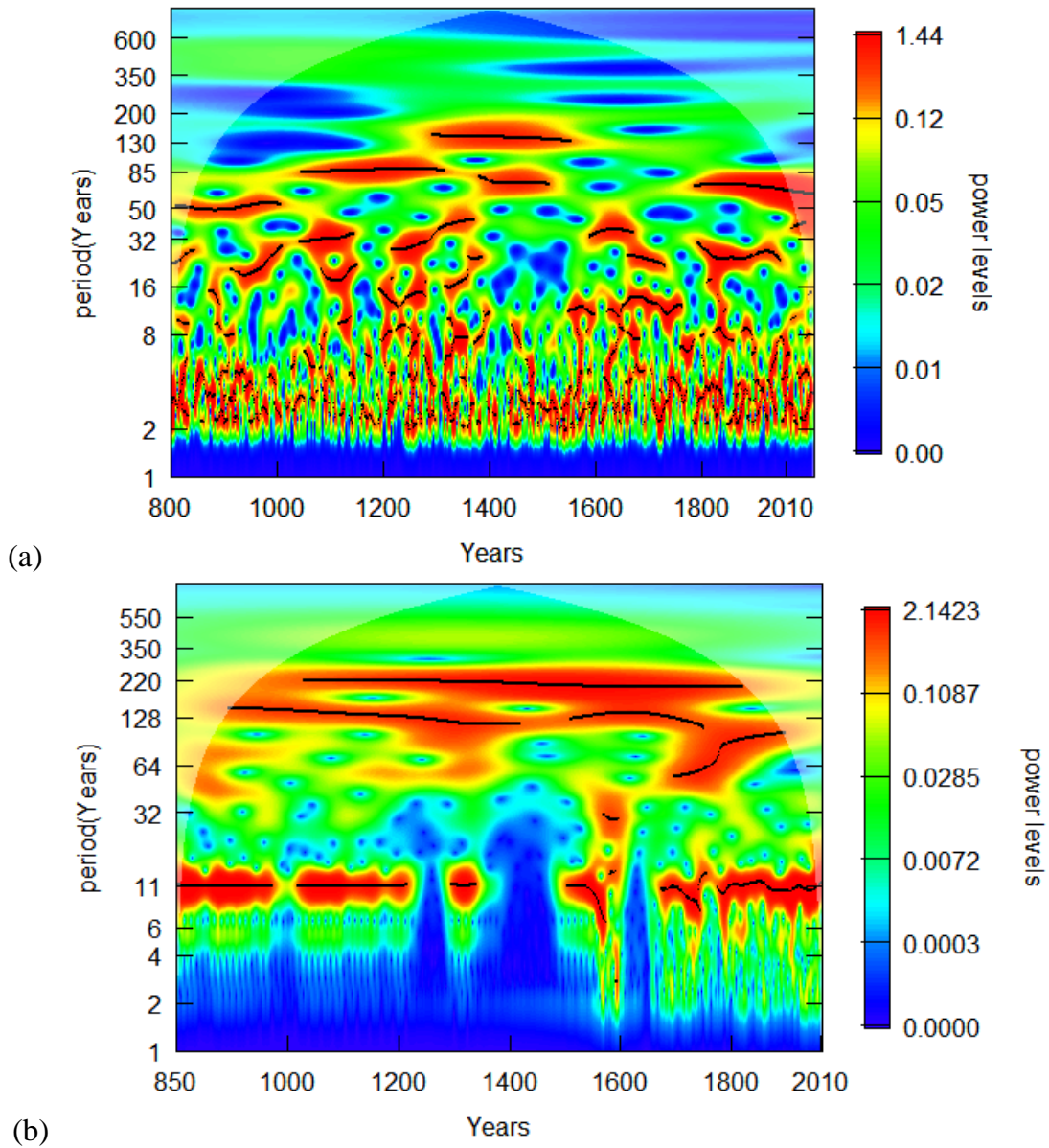

(a)

(b)

Figure 2. Wavelet spectrums of (a) AMO and (b) TSI

Outside the COI, statistically significant regions at the 0.05 level are enclosed by a thick black contour. On the *y*-axis, each series is decomposed into 8 – time scales, where the shortest one (i.e., 1/2 – 1 years) denotes the highest frequency band and the longest one (i.e., 256 – 512 years) denotes the lowest frequency band.

From the obtained CWT results shown in fig. 2 one can determine all oscillations with small amplitudes, and can detect time intervals with significant amplitude of oscillations as well as their frequency variations.

Second, the wavelet spectrum of the AMO data reveals minor centennial Suess - de Vries cycles with periods from 195– to 235 – year; variable pieces of Gleisberg cycles with periods 70 – 130 years; a mode of solar rotation with period 50-60 years; and variable solar harmonics with periods 20 – 50 years, while of the TSI data shows almost constant Suess - de Vries cycles with a period from 195 – to 235 – year; variable Gleisberg cycles; and constant period of 11-year cycles for the first half of time series.

### 5.3 WAVELET COHERENCE OF TSI AND AMO

In order to search for statistically relevant correlation between the periodicities found in the AMO and TSI data, we performed a cross wavelet and wavelet coherence analyses. The CWT is used to analyze the AMO and TSI signals. Similar to a common power spectrum, the CWT expands a time series into frequency space, but without losing the time information and allows analyzing time-localized oscillations. To find a relationship between the AMO and the TSI, we use wavelet coherence (WC) which compares two CWTs and finds locally phase-locked behavior of two signals. By using Monte Carlo simulations, the statistical significance level of the WC is tested. For each scale of the CWT, the significance level is estimated by calculating the wavelet coherence with each pair of datasets (Grinsted et al., 2004).

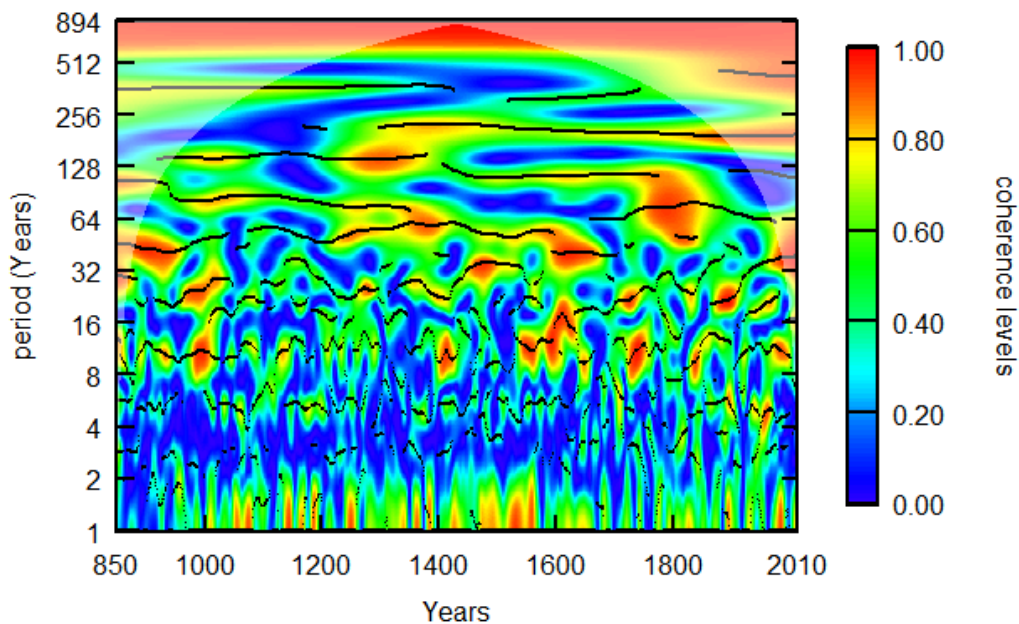

**Figure 3.** Wavelet coherence spectrum between the two TSI and AMO time series

The relationship between the AMO and TSI annual time series over the period 850–2011 years was calculated by wavelet coherence. The absolute coefficients are shown in Fig. 3. As we known from (Chapanov, 2021) between the two TSI and AMO time series there are correlation periodicities shown in Table 1. In three dimensions, the results of wavelet coherence in fig.3 show significant coherence between the AMO and TSI. It is used to quantify the degree of association of two nonstationary series in the time – frequency domain.

## 6. CONCLUSIONS

This paper mainly focuses on the relationship between sunspot number and Atlantic multidecadal oscillation.

First, the results show that in contrast to the Fourier analysis, which determines all oscillations with small amplitudes, the CWT detects time intervals with significant amplitude of oscillations and their frequency variations.

Second, the obtained CWT of the AMO data reveals minor cycles; variable pieces of Gleisberg cycles; a mode of solar rotation; and variable solar harmonics, almost constant cycles for the TSI data and constant period of 11 – year cycles for the first half of time series.

Finally, the correlation between the AMO-TSI expressed by coherence is significant for oscillations close to 7-th harmonic of 2300-year Hallstatt solar cycle with 300 – year period; accelerated cycles of Suess - de Vries and accelerated Gleisberg cycles.

One important conclusion is that applications of both Fourier and wavelet analyses can significantly improve interdisciplinary research.

## ACKNOWLEDGEMENTS

The study is supported by the National Science Fund of Bulgaria, Contract KP-06-N34/1 /30-09-2020 "Natural and anthropogenic factors of climate change – analyzes of global and local periodical components and long-term forecasts"

## DATA AVAILABLE

All the data used in this work are available freely in the internet. The TSI data used in this study are available online at
*https://lasp.colorado.edu/lisird/data/nrl2_tsi_P1Y/*
and the AMO data – at
*https://www.ncei.noaa.gov/pub/data/paleo/reconstructions/wang2017/wang2017amv-amo.txt*